# DeepSeek in Healthcare: A Survey of Capabilities, Risks, and Clinical Applications of Open-Source Large Language Models

Jiancheng Ye, PhD[1*], Sophie Bronstein, MD[1], Jiarui Hai, MS[2], Malak Abu Hashish, MS[3]

1 Weill Cornell Medicine, Cornell University, New York, New York, USA;

2 Department of Engineering, Johns Hopkins University, Baltimore, Maryland, USA;

3 Touro University College of Osteopathic Medicine, Middletown, New York, USA

**Corresponding author:**
Jiancheng Ye, PhD
Weill Cornell Medicine, Cornell University, New York, NY, USA
jiancheng.ye@u.northwestern.edu

**ABSTRACT**

DeepSeek-R1 is a cutting-edge open-source large language model (LLM) developed by DeepSeek, showcasing advanced reasoning capabilities through a hybrid architecture that integrates mixture of experts (MoE), chain of thought (CoT) reasoning, and reinforcement learning. Released under the permissive MIT license, DeepSeek-R1 offers a transparent and cost-effective alternative to proprietary models like GPT-4o and Claude-3 Opus; it excels in structured problem-solving domains such as mathematics, healthcare diagnostics, code generation, and pharmaceutical research. The model demonstrates competitive performance on benchmarks like the United States Medical Licensing Examination (USMLE) and American Invitational Mathematics Examination (AIME), with strong results in pediatric and ophthalmologic clinical decision support tasks. Its architecture enables efficient inference while preserving reasoning depth, making it suitable for deployment in resource-constrained settings. However, DeepSeek-R1 also exhibits increased vulnerability to bias, misinformation, adversarial manipulation, and safety failures - especially in multilingual and ethically sensitive contexts. This survey highlights the model's strengths, including interpretability, scalability, and adaptability, alongside its limitations in general language fluency and safety alignment. Future research priorities include improving bias mitigation, natural language comprehension, domain-specific validation, and regulatory compliance. Overall, DeepSeek-R1 represents a major advance in open, scalable AI, underscoring the need for collaborative governance to ensure responsible and equitable deployment.

**INTRODUCTION**

**The rise of AI and generative models in health and technology**

Artificial Intelligence (AI) has undergone transformative growth in recent years, profoundly reshaping numerous fields including language processing, automation, and complex decision-making. At its core, AI refers to the simulation of human intelligence by machines, enabling them to perform tasks such as speech recognition, natural language understanding, visual perception, and predictive analytics.[1] Among the most influential developments in this space is Generative AI—a branch of AI that focuses on producing original content such as text, images, code, and even audio or video.[2] These systems are predominantly driven by deep learning and large-scale data modeling.

One of the recent remarkable advancements in the Generative AI domain is the emergence of DeepSeek-R1, a large language model (LLM) developed by the Chinese company DeepSeek.[3] Released in January 2025, DeepSeek-R1 has garnered significant attention for its advanced reasoning capabilities and performance across a wide array of complex tasks. In benchmarking evaluations, it has demonstrated results competitive with, and in some domains superior to, models like OpenAI's GPT-4o and GPT-o1[4]. This has positioned DeepSeek-R1 as a notable advancement not only in LLM capability but also in the global AI development race.

**DeepSeek-R1: a paradigm shift in LLM development**

What sets DeepSeek-R1 apart from conventional LLMs is its novel training architecture. While most existing models rely heavily on supervised learning using static, pre-curated datasets, DeepSeek-R1 incorporates reinforcement learning (RL) into its framework—enabling it to learn iteratively through dynamic feedback and interaction. This hybrid approach mimics certain aspects of human learning, allowing the model to refine its behavior over time and adapt to more complex reasoning tasks. As a result, DeepSeek-R1 exhibits marked improvements in logical reasoning, mathematical problem-solving, and code generation, surpassing several peers in structured cognitive tasks[5]. Its architecture supports advanced capabilities such as stepwise problem decomposition and chain-of-thought (CoT) reasoning, which mirror the human approach to solving multi-step problems. These features enable the model to perform well in domains that require multi-layered logic and abstraction. However, DeepSeek-R1's innovation also introduces new challenges. Security researchers have uncovered vulnerabilities in its CoT reasoning pipeline—an essential component intended to replicate human-like step-by-step thought processes[6]. Techniques such as Hijacking Chain-of-Thought (H-CoT) have been used to manipulate the model's outputs by exploiting this reasoning structure, highlighting unresolved concerns related to AI alignment, safety, and trustworthiness[7]. These issues are particularly critical in high-stakes domains like healthcare, where model errors or manipulations can have serious consequences.

  Despite these concerns, DeepSeek-R1 demonstrates strong cross-domain potential. In healthcare settings, early evaluations suggest that the model can provide diagnostic support in pediatric medicine at a level comparable to ChatGPT-o1. Similarly, in the financial industry, DeepSeek-R1 has been employed to support high-accuracy predictive modeling while maintaining relatively low computational overhead—an important factor for broader adoption in resource-constrained environments[8]. Its mathematical rigor and programmatic fluency further differentiate it from other open models. DeepSeek-R1 has shown remarkable capability in formal theorem proving, algorithm design, and multi-turn dialogue systems. These competencies position it as a valuable tool for research, education, and potentially clinical reasoning, although real-world deployment in areas like drug discovery, personalized medicine, and clinical decision-making remains in the experimental phase[9]. As with all AI models entering healthcare, concerns around safety, generalizability, liability, and ethical use must be carefully addressed before widespread implementation[10].

**Openness and democratization of AI**

A defining feature of DeepSeek-R1 is its open-weight release under the permissive MIT license. This level of transparency marks a strategic departure from the closed, proprietary approach taken by companies such as OpenAI, Google DeepMind, and Anthropic. Open-source release not only enables academic researchers, startups, and developers to audit, modify, and build upon the model, but also promotes transparency, reproducibility, and collaborative innovation in the broader AI community. This openness has significant implications for the democratization of AI, particularly in regions or institutions with limited resources. By lowering the barrier to entry, open-source models like DeepSeek-R1 allow for more equitable participation in cutting-edge AI development and application. In healthcare, where access to high-quality AI tools is often constrained by financial and infrastructural limitations, open models could play a transformative role in improving clinical support and public health surveillance.

The open-source nature of DeepSeek-R1 is also a double-edged sword. While it enables innovation, it also exposes the model to adversarial fine-tuning, prompt injection, and alignment manipulation. Studies have shown that malicious actors can exploit its reasoning mechanisms—particularly those based on CoT—to steer outputs toward harmful, biased, or misleading content[11]. Furthermore, its advanced reasoning mechanisms, including CoT reasoning, while enhancing performance, simultaneously increase vulnerability to prompt injection and reward hacking. These concerns underscore the need for robust alignment strategies, red-teaming protocols, and ongoing community oversight to ensure responsible use. As the global AI landscape evolves, DeepSeek-R1 represents both a technological milestone and a case study in the challenges of deploying high-capability, open-source LLMs in sensitive domains like healthcare. Its emergence calls for a nuanced understanding of the technical, ethical, and regulatory dimensions of generative AI—and for collaborative efforts across sectors to harness its potential responsibly.

**TRAINING AND ARCHITECTURE**

**Multi-stage training: from pretraining to self-reflection**

DeepSeek-R1 is trained through a sophisticated, multi-stage process that goes well beyond conventional supervised fine-tuning. At the core of this development pipeline is a layered learning approach that emphasizes not just content comprehension but also reasoning optimization and self-evaluation. The training begins with pretraining on DeepSeek-V3, leveraging a cost-efficient Mixture of Experts (MoE) architecture. This architecture enables the model to selectively activate subsets of specialized neural

"experts" depending on the input, thereby balancing scalability and performance. During this phase, DeepSeek-R1 is exposed to massive amounts of unlabeled data across multiple domains and languages, enabling it to form a robust foundational understanding of linguistic and logical patterns.

Following pretraining, the model advances to Supervised Fine-Tuning (SFT). In this stage, DeepSeek-R1 is refined using curated datasets tailored to specific downstream tasks. SFT aligns the model's output with desired user intents, enhancing its reliability in practical applications such as question-answering, summarization, and code generation. Next, the model undergoes Reinforcement Learning from Human Feedback (RLHF), a widely used strategy for aligning AI behavior with human values. DeepSeek-R1 employs a novel algorithm called Group Relative Policy Optimization (GRPO), which guides the model through a reward-based learning loop. This trial-and-error framework introduces a dynamic learning mechanism in which the model tests different reasoning pathways, receives reward signals based on output quality, and progressively refines its internal logic. A particularly novel element of this process is self-reflection, where the model critiques and revises its own reasoning strategies—producing emergent "aha moments" that simulate human-like learning and insight development [5].

**Figure 1** presents the workflow of DeepSeek-R1, encompassing both the training process and the model's output generation. The training pipeline begins with data collection using CoT examples, followed by reasoning-oriented reinforcement learning and supervised fine-tuning. Multiple reward signals—targeting accuracy, formatting, and linguistic coherence—are integrated to better align the model's behavior with human preferences. The output generation workflow demonstrates how the model encodes input queries, performs internal reasoning through CoT, synthesizes structured responses, and generates final outputs that are accurate, coherent, and contextually appropriate. Together, these processes underscore DeepSeek-R1's emphasis on structured reasoning and interpretability.

**Architecture innovations: MoE and multihead latent attention**

At inference time, DeepSeek-R1's efficiency is underpinned by its MoE framework. Unlike dense transformer models that activate the entire model for each input, MoE activates only a subset of its neural experts, selected based on the input context. This targeted activation significantly reduces computational load without sacrificing output quality. Furthermore, the use of Multihead Latent Attention (MLA) enhances the model's ability to integrate long-range dependencies across tokens, improving contextual understanding while maintaining computational efficiency [4].

Empirical studies reveal that this architecture yields superior performance in tasks requiring specialized knowledge, such as cross-lingual translation, short-video script generation, and domain-specific dialogue generation [12]. However, long CoT reasoning processes—while improving accuracy—introduce latency in processing time. This can be problematic in real-time or high-throughput environments, where rapid response generation is critical [13].

**Advanced reasoning and symbolic computation**

DeepSeek-R1 distinguishes itself in symbolic reasoning and mathematical problem-solving through a token-rich, multi-step generation process. Unlike models that prioritize brevity and speed, DeepSeek-R1 takes a more elaborate route—generating more tokens per response in order to simulate human-like logical deduction. This method proves highly effective in formal reasoning tasks such as theorem proving, algorithm design, and multi-turn problem decomposition. However, it also increases computational costs in environments requiring rapid inference[14].

A defining strength of DeepSeek-R1 is its CoT reasoning capability, which allows it to break down complex problems into structured, interpretable steps. During RLHF training, CoT is reinforced through a three-phase loop: 1) Trial-and-error learning: The model iteratively attempts solutions to challenging tasks[5]; 2) Reward feedback: Successful strategies are positively reinforced; and 3) Self-reflection and optimization: The model analyzes its own problem-solving methods to discover patterns or shortcuts that lead to deeper understanding[4, 5].

This multi-tiered reasoning pipeline has translated into measurable improvements in domain-specific tasks. For instance, DeepSeek-R1 has performed competitively on the United States Medical Licensing Examination (USMLE), showing comparable accuracy to GPT-4o despite using fewer training resources. In ophthalmology diagnostic tasks, the model has demonstrated similar accuracy to proprietary models like GPT-o1, while incurring significantly lower computational costs [4]. These results highlight DeepSeek-R1's potential as a cost-effective alternative for specialized domains, especially in resource-constrained environments such as rural healthcare settings or developing countries. The model's design shows promise for democratizing AI access while maintaining performance integrity in domain-critical applications.

Nonetheless, significant hurdles remain. RLHF continues to suffer from alignment issues such as reward gaming, where outputs are tailored to exploit reward functions rather than ensure safety or factuality. Language inconsistency and token overproduction are also recurring concerns, especially in applications requiring precision and interpretability, such as medical diagnostics or legal analysis. Moreover, the generalizability of DeepSeek-R1 across multiple domains is still limited. While the model excels in logic-heavy tasks, its performance tends to degrade in less-structured, ambiguous environments. This restricts its adaptability for real-world, multi-domain deployment in areas like clinical informatics,[15] electronic health record summarization,[16] or complex biomedical research pipelines[17].

**COMPARISONS WITH OTHER MODELS**

**Table 1** demonstrates the comparative analyses of key characteristics of DeepSeek-R1 and other recent LLMs.

**1. DeepSeek-R1 vs Open-Source LLMs**

**1.1 Architectural differences**

DeepSeek-R1 distinguishes itself from other prominent open-source LLMs such as LLaMA and Qwen through its advanced MoE architecture[5]. Unlike traditional transformer models, which employ a single, dense network to process all inputs uniformly, DeepSeek-R1 activates a sparse subset of expert subnetworks tailored to specific input features. This modular structure enables more efficient computation by routing each input through only the most relevant experts, reducing unnecessary parameter activation and memory usage. This architecture grants DeepSeek-R1 a strategic advantage, particularly for tasks that require specialized forms of reasoning, such as mathematical problem-solving and code generation. By leveraging domain-specific experts, the model can isolate and address different facets of a complex task more effectively than monolithic architectures like those used in LLaMA and Qwen. Additionally, DeepSeek-R1 integrates reinforcement learning (RL) with a Gradient-Based Policy Optimization (GRPO) mechanism during training, enhancing its ability to refine problem-solving strategies iteratively. This contrasts with the conventional use of supervised fine-tuning (SFT) in LLaMA and Qwen, which may limit adaptability across tasks requiring multi-step reasoning.

## 1.2 Performance benchmarking

DeepSeek-R1 demonstrates leading-edge performance in structured domains. On the American Invitational Mathematics Examination (AIME) 2024 benchmark for high school-level mathematical reasoning, it achieved an impressive accuracy of 86.7%, and on Codeforces, it ranked in the 96.3rd percentile, showcasing its proficiency in logic-based and algorithmic coding tasks[14]. These achievements are underpinned by its CoT reasoning mechanism, which decomposes complex problems into a sequence of smaller, interpretable steps—enhancing not only accuracy but also transparency in decision-making. However, in more general natural language understanding (NLU) tasks, DeepSeek-R1 is slightly less competitive. It scored 71.5% on the GPQA Diamond dataset, indicating solid performance but trailing behind top open-source models more optimized for linguistic coherence and conversational fluency[18]. This trade-off reflects DeepSeek-R1's intentional prioritization of high-reasoning applications over general-purpose NLP.

## 1.3 Accessibility and transparency

DeepSeek-R1, released under the permissive MIT license, exemplifies a commitment to transparency and community empowerment[4]. Like LLaMA and Qwen, its model weights and architecture are openly available. However, its documentation and modular MoE framework are particularly developer-friendly, allowing advanced users to isolate, modify, and repurpose specific expert modules for custom applications. This design fosters reproducibility and insight into the model's internal mechanics—benefits often absent in closed ecosystems. Developers can trace output decisions back to specific experts, offering interpretability and diagnostic capabilities that are valuable for both research and real-world deployments.

## 1.4 Bias and Safety

The openness of DeepSeek-R1 introduces important safety considerations. Its capacity for fine-tuning and adaptability through CoT reasoning increases susceptibility to adversarial manipulation. Research has shown that attackers can exploit these features to fine-tune the model toward generating biased, misleading, or even harmful outputs[11]. While LLaMA and Qwen face similar challenges due to their transformer architectures, DeepSeek-R1's modular and CoT-centric structure may amplify the risk unless robust safety protocols are applied. However, the active participation of the open-source community provides an avenue for rapid development and deployment of bias-mitigation strategies—an advantage not readily available in proprietary models.

## 1.5 Cost efficiency and scalability

DeepSeek-R1's selective activation of experts significantly reduces computational costs during inference. By engaging only the relevant submodules, it avoids the heavy compute burden of fully dense transformer layers. This makes the model particularly attractive for deployment in environments with limited hardware or budget constraints[5]. Though its reinforcement learning training phase is initially resource-intensive, it yields long-term efficiency by reducing the need for continual re-training. Compared to SFT-based alternatives that may require frequent updates to maintain performance, DeepSeek-R1 offers an efficient and sustainable lifecycle cost profile.

## 2. DeepSeek-R1 vs Closed-Source LLMs

### 2.1 Architectural differences

DeepSeek-R1's architecture offers a rare glimpse into the potential of modular, open-source AI. Unlike proprietary models such as GPT-4o, GPT-o1, Claude-3 Opus, and Mixtral, whose internal structures are either partially or entirely opaque, DeepSeek-R1 provides full visibility into its components[5]. DeepSeek-R1's MoE architecture allows for optimized, task-specific computation, selectively activating experts based on input complexity and domain. For instance, Mixtral—developed by Mistral AI—is one of the few closed-source models that also uses MoE, though it activates fewer experts per query than DeepSeek-R1. Meanwhile, GPT-4o integrates multimodal capabilities (text, image, audio) but relies on dense transformer layers. Claude-3 Opus focuses heavily on ethical alignment and robust reasoning, but details of its internal mechanics remain undisclosed[19]. DeepSeek-R1's open MoE design, paired with reinforcement learning and GRPO, provides a level of flexibility and transparency unmatched by closed-source systems. Researchers can audit, replicate, and refine the model—contributing to scientific progress and democratizing access to advanced AI[5].

## 2.2 Performance benchmarking

Despite its open-source nature, DeepSeek-R1 rivals closed-source models in several key domains. Its 86.7% accuracy on AIME and 96.3rd percentile in Codeforces demonstrate its top-tier capabilities in mathematical reasoning and code generation—often on par with GPT-o1 and GPT-4o[17]. In a task involving ophthalmology-based questions, DeepSeek-R1 matched GPT-o1's 82% accuracy but at a fraction of the computational cost. While its conversational NLU lags slightly behind GPT-4o and Claude-3 Opus (both scoring above 75% on general NLP benchmarks), DeepSeek-R1's interpretable CoT approach provides more granular reasoning, particularly useful in structured decision-making scenarios[18]. However, the CoT mechanism also leads to higher token usage, reducing response efficiency in latency-sensitive applications like real-time chat. Closed-source models may offer a performance edge in such settings, but DeepSeek-R1's adaptability allows developers to refine and optimize it for specific workflows[19].

**2.3 Accessibility and transparency**

One of DeepSeek-R1's most valuable differentiators is its unrestricted access. Under the MIT license, developers can deploy the model locally, integrate it into proprietary workflows, and customize it freely—all without licensing costs or API usage fees[4]. By making its model weights publicly accessible, DeepSeek-R1 allows a broader range of users to experiment with, adapt, and deploy this AI system without the steep financial or licensing barriers that typically accompany closed-source models. In contrast, accessing models like GPT-4o or Claude-3 Opus generally requires paid API subscriptions, institutional partnerships, or restricted enterprise agreements, limiting their use to well-funded or commercial environments. Beyond access, DeepSeek-R1's transparency allows for educational and technical exploration, as developers can study and build on its MoE architecture, GRPO, and CoT reasoning workflows, enabling in-depth learning and experimentation that is not possible with the black-box design of proprietary systems. This openness fosters a collaborative, decentralized innovation model, where contributors can refine and extend the system[5].

**2.4 Bias and Safety**

DeepSeek-R1's transparency is a double-edged sword. While it facilitates community-led efforts to identify and correct biases, it also exposes the model to fine-tuning attacks[11]. The CoT framework enhances performance in structured tasks, but also increases the model's vulnerability to adversarial prompting and manipulation, making bias mitigation a critical concern. In contrast, closed-source models like GPT-4o and Claude-3 Opus conceal their alignment methodologies, making it difficult to assess their response against similar threats[20]. Claude-3 Opus, in particular, is known for its strong emphasis on safety and ethical AI, likely employing RLHF and constitutional AI principles to minimize bias. However, its proprietary nature makes any biases or vulnerabilities hidden from the public. DeepSeek-R1's transparency provides a unique advantage—allowing for refinements to safeguard against harmful outputs. This trade-off between openness and controlled safety measures highlights a core juxtaposition within AI development, as DeepSeek-R1 prioritizes collective improvement and adaptability, while proprietary models lean toward centralized oversight and restrictive access to alignment mechanisms.

## 2.5 Cost efficiency and scalability

The MoE architecture of DeepSeek-R1 offers significant cost advantages. For example, in ophthalmology tasks, DeepSeek-R1 performed comparably to OpenAI's o1 model, but with inference costs almost 15 times lower[17]. This efficiency stems from its selective activation of expert modules, reducing unnecessary computation and energy usage compared to fully dense models like GPT-4o. While the internal architectures of proprietary models remain unclear, their API-based access, high token usage, and lack of deployability introduce scalability and cost challenges for organizations. In contrast, DeepSeek-R1's open-source nature allows individuals and institutions to deploy the model locally, minimizing dependency on expensive cloud services. This can lead to substantial long-term savings and greater control over data and infrastructure. Although the use of CoT reasoning may increase token consumption during complex tasks, DeepSeek-R1 remains highly suited to scenarios requiring deep reasoning. Its open architecture also enables ongoing, community-driven optimizations, enhancing both performance and cost-effectiveness over time[21].

**CLINICAL APPLICATIONS**

DeepSeek-R1 has been deployed across diverse healthcare settings, where its advanced reasoning and natural language capabilities support a broad range of clinical, educational, and research applications. The model's strengths in clinical decision-making, diagnostic support, medical education, pharmaceutical research, and patient communication highlight its potential to transform modern medicine.

**Table 2** summarizes the selected studies evaluating DeepSeek-R1 across various domains, especially healthcare. DeepSeek-R1 demonstrated strong capabilities in diagnosis, drug development, and decision support, with performance comparable to or exceeding other LLMs in tasks like USMLE-style questions, thoracic oncology, and ophthalmology. However, performance varied by language and specialty, with ChatGPT often outperforming in English-based or pediatric tasks. In AI safety, there are concerns about DeepSeek-R1's vulnerability to generating unsafe content, particularly in adversarial or multilingual settings. The model showed more harmful outputs than o3-mini and exhibited limitations in alignment, especially in Chinese contexts. Researchers emphasized the need for hybrid alignment strategies (e.g., RL + SFT) and domain-specific safety tuning. Technological assessments revealed DeepSeek-R1's strong reasoning and multilingual capabilities but also highlighted efficiency trade-offs due to high token usage. In academic content creation and video generation, it performed well but faced originality and detection issues. Overall, DeepSeek-R1 is cost-effective and promising, especially for resource-constrained applications, but its safety, reasoning reliability, and domain-specific performance require further improvement. The studies collectively underscore the importance of alignment, transparency, and optimization for broader deployment.

Below, we detail the scope and impact of its current implementations.

**Clinical decision support in pediatrics**

In pediatric medicine, DeepSeek-R1 has demonstrated robust performance in supporting clinical decisions. Researchers have leveraged the model's CoT reasoning capabilities to dissect complex patient cases, synthesize symptoms, and propose diagnoses and treatment options for both common and rare pediatric conditions. By breaking down clinical queries into logical components, the model helps identify subtle risk factors and patterns that may be overlooked by human providers, particularly in time-sensitive scenarios[22]. This structured approach has reduced diagnostic errors and improved care delivery in critical care contexts. In a comparative evaluation using the MedQA pediatric dataset, DeepSeek-R1 achieved an 87.0% diagnostic accuracy—just below ChatGPT-o1's 92.8%—but stood out due to its open-source architecture and reflective reasoning capabilities. These features have made it especially valuable in resource-constrained environments, where access to commercial LLMs may be limited or cost-prohibitive[22].

**Advancing ophthalmic diagnosis and efficiency**

In ophthalmology, DeepSeek-R1 has shown impressive promise by enhancing diagnostic accuracy and reducing operational costs. The model's ability to integrate textual clinical data with imaging inputs—such as retinal scans—supports more comprehensive assessments for diseases like glaucoma and macular degeneration[17]. A comparative analysis involving 300 ophthalmology cases revealed that DeepSeek-R1 achieved an accuracy of 82.0%, comparable to OpenAI's o1 model, while incurring a 15-fold lower cost per query [17]. Beyond accuracy, its affordability and efficiency make DeepSeek-R1 well-suited for deployment in large-scale or economically constrained healthcare settings. Its use in semi-automated diagnostic workflows also helps reduce the burden on ophthalmologists, allowing them to focus on high-complexity cases and direct patient care[17].

**Medical education and examination performance**

One of DeepSeek-R1's most impactful applications is in medical education. The model excels in educational environments by offering high-quality explanations, step-by-step clinical reasoning, and customized feedback. It consistently surpasses expectations on medical licensure assessments such as the United States Medical Licensing Examination (USMLE), showcasing its ability to parse multi-step clinical questions and provide evidence-based justifications[23]. DeepSeek-R1 was also a top-tier performer in the MedXpertQA benchmark—designed to test real-world clinical reasoning across 11 body systems and 17 medical specialties—further solidifying its role as a valuable educational tool[24]. In virtual learning environments, it helps medical trainees hone their decision-making skills by simulating diagnostic reasoning in real time.

**Patient education and health literacy**

In addition to professional education, DeepSeek-R1 contributes to patient-centered care through the generation of accessible and easy-to-read educational content. In a recent comparison of scoliosis-related materials, the model produced outputs with a Flesch-Kincaid Grade Level of 6.2 and a Flesch Reading Ease score of 64.5—outperforming ChatGPT-o1 on both metrics[25]. This capacity to tailor content to a broad audience makes it a valuable resource for promoting health literacy, particularly in populations with limited medical knowledge.

**Secure, no-code clinical applications**

The model's design allows for seamless integration into no-code, privacy-preserving healthcare applications. In nephrology, for example, clinicians have deployed DeepSeek-R1 locally without relying on cloud infrastructure, ensuring compliance with data privacy standards while maintaining model functionality[26]. This deployment model is especially critical for institutions that handle sensitive health data or operate under strict regulatory constraints.

**Pharmaceutical research and drug interaction prediction**

Beyond clinical applications, DeepSeek-R1 is gaining recognition in pharmaceutical research, particularly in predicting drug-drug interactions (DDIs). By leveraging its extensive understanding of molecular biology, pharmacodynamics, and genomics, the model supports early-stage identification of adverse interactions—reducing the risks associated with polypharmacy and enhancing drug development pipelines.[27] In a benchmark study involving 18 LLMs on DDI prediction using DrugBank data, DeepSeek-R1 did not outperform the top model, Phi-3.5, but demonstrated competitive performance and strong potential when fine-tuned[28]. Its ability to analyze complex molecular and genetic structures gives it a unique advantage in addressing safety and efficacy concerns in combination therapies, particularly for aging and chronically ill populations.

**Barriers to adoption and public trust**

Despite its promising applications, real-world deployment of DeepSeek-R1 in healthcare systems faces challenges. Concerns about model transparency, ethical data use, and algorithmic bias remain key barriers to widespread adoption. These issues are particularly significant in high-stakes settings, where even small errors can have life-altering consequences[29]. Recent cross-national surveys in the United States, United Kingdom, and India have shown that public trust in DeepSeek-R1 is influenced by perceived ease of use and perceived utility, but also hampered by concerns over potential risks[30]. For broader acceptance, developers and healthcare organizations must prioritize transparency around training data, implement rigorous validation protocols, and engage in continuous risk assessment to mitigate bias and improve model reliability.

**STRENGTHS AND LIMITATIONS OF DEEPSEEK-R1**

DeepSeek-R1 emerges as a versatile LLM, exhibiting exceptional effectiveness across domains such as medicine, finance, mathematics, and computer science. One of its core strengths lies in its ability to employ CoT reasoning, allowing it to decompose complex tasks into logical, manageable steps. This capacity enhances its performance in diagnostic decision-making, mathematical computation, and code generation. Furthermore, its open-source architecture and cost-effectiveness have made it particularly appealing for research and implementation in resource-constrained settings.[31] However, despite its strengths, DeepSeek-R1 is not without its drawbacks. Like other LLMs, it presents risks of bias, hallucination, and misuse—challenges that are magnified by its open-source accessibility and potential for adversarial fine-tuning. Moreover, the model occasionally struggles with tasks involving nuanced natural language understanding and ethical decision-making, which are critical in sensitive domains like healthcare and policy.[32]

**Strengths**

***Superior Performance in Logic and Quantitative Reasoning***

DeepSeek-R1 consistently outperforms many competing models—including Claude-3 Opus and Mixtral—on complex quantitative tasks. This includes standardized tests like the AIME, advanced coding challenges, and intricate logical reasoning problems. The combination of CoT reasoning and reinforcement learning techniques enhances its capacity to produce systematic, high-accuracy outputs.

*Continuous and adaptive learning potential*

Leveraging reinforcement learning (RL), DeepSeek-R1 supports incremental knowledge acquisition through trial and error. This adaptability allows the model to refine its logic and incorporate new scientific or technical developments over time. In clinical practice, for instance, this trait enables rapid updates in response to evolving medical guidelines or research discoveries, offering a dynamic decision-support system[29].

*High Cost-Efficiency for Inference*

One of DeepSeek-R1's most compelling features is its exceptionally low inference cost—reportedly 27 times cheaper than OpenAI's o1 model. This affordability makes it a viable solution for startups, academic institutions, and healthcare systems aiming to harness advanced AI capabilities without incurring prohibitive operational expenses[8].

*Open-source and customizable architecture*

Unlike many proprietary LLMs, DeepSeek-R1 provides openly accessible model weights. This transparency facilitates fine-tuning, integration into domain-specific pipelines, and collaborative innovation. The open-source nature fosters community-driven experimentation and reproducibility, advancing the field of responsible AI[4, 33].

*Structured iterative reasoning*

The model's CoT reasoning, when combined with iterative RL optimization, enhances its ability to tackle multi-step problems. This structured problem-solving methodology not only improves output quality but also allows domain experts to trace and validate the model's reasoning processes, boosting transparency and trust[23].

*Scalability and modularity*

DeepSeek-R1 can be distilled into smaller, fine-tuned versions without significant performance degradation[34]. This makes it adaptable for deployment across a range of computing environments, from high-performance clusters to local edge devices—enabling both cloud-based and offline applications[33].

*Viability in clinical and low-resource settings*

Although slightly less accurate than top-tier models like GPT-4o in clinical benchmarks, DeepSeek-R1 offers a cost-effective alternative for diagnostic and decision-support tasks in under-resourced environments. It balances affordability with strong baseline performance, making it well-suited for global health initiatives and telemedicine use cases[22].

*Broad domain versatility*

The model demonstrates robust cross-domain competence, from scientific research and software engineering to health informatics. In particular, its strong performance in clinical informatics and decision-making—when paired with its low operational cost—makes it an appealing candidate for integrative AI solutions in healthcare[8, 22, 23, 28].

*Offline usability and enhanced privacy*

Unlike many commercial models, DeepSeek-R1 supports offline deployment, which minimizes data transfer risks and enhances compliance with data privacy requirements. This feature is especially beneficial for institutions dealing with sensitive or regulated data, such as hospitals or financial organizations[29].

**Limitations**

*Security risks and adversarial fine-tuning*

The openness of DeepSeek-R1 introduces significant vulnerability to adversarial fine-tuning. Malicious actors could potentially exploit this flexibility to override safety constraints, leading to outputs that are harmful or misaligned with ethical norms. Even in offline settings, unclear data retention policies present additional risks, particularly for applications involving personal health or financial data [11, 35].

*increased propensity for bias and misinformation*

Studies suggest that DeepSeek-R1 is substantially more likely to produce biased or misleading outputs compared to its proprietary peers. It is reportedly three times more prone to bias than Claude-3 Opus and four times more likely to propagate misinformation than GPT-4o. This vulnerability is especially critical in policy or healthcare contexts, where such errors can have tangible negative consequences[29].

*Unpredictability due to over-customization*

While the model's open-source nature offers flexibility, inconsistent or poorly executed customizations can result in unstable or unsafe behaviors. In high-stakes domains like medicine or law, this inconsistency may undermine trust and reliability[29].

*Susceptibility to harmful or toxic content*

Red-teaming efforts have demonstrated that DeepSeek-R1 may generate toxic, offensive, or dangerous content at higher rates than models like OpenAI's o1. Examples include erroneous code recommendations and unsafe advice concerning chemical synthesis. These findings underscore the need for rigorous oversight and access controls[20].

*Computational overhead in complex reasoning*

Although DeepSeek-R1 is efficient in terms of basic inference, its token-intensive reasoning style can drive up computational demands in large-scale or real-time applications. As a result, organizations running many concurrent queries or working under tight latency constraints may find its performance suboptimal for such use cases[14].

*Regulatory and Legal Ambiguity*

There remains uncertainty regarding DeepSeek-R1's adherence to global privacy and safety standards, including GDPR, HIPAA, and other compliance frameworks[29].

In fields like finance and healthcare, this regulatory ambiguity poses substantial barriers to deployment and scalability[8, 17].

*Subpar performance in language understanding and creativity*

While DeepSeek-R1 excels in structured problem-solving, it underperforms in tasks that require nuanced language comprehension, emotional intelligence, or creativity. Compared to GPT-4o or Claude-3 Opus, it struggles with casual conversation, literary generation, and subtleties in tone or intent—limiting its usefulness for certain user-facing applications[33].

*Monitoring and ethical oversight*

Due to its modifiable architecture and evolving dataset, continuous evaluation is critical. Ensuring alignment with ethical standards, accuracy benchmarks, and security protocols demands persistent oversight, particularly as the model is adopted in more sensitive or mission-critical environments[19, 23]

**ETHICAL CONSIDERATIONS**

The advanced capabilities of DeepSeek-R1 open up significant opportunities for innovation across domains. However, these capabilities also give rise to serious ethical, social, and legal challenges. **Figure 2** demonstrates the ethical considerations of DeepSeek-R1. Key issues include bias, misinformation, misuse, privacy risks, and regulatory uncertainty. It is imperative to proactively address these concerns to ensure that the deployment of such powerful models serves the public good, avoids reinforcing existing disparities, and prevents harmful applications.

**Algorithmic bias and misinformation**

Despite its impressive technical performance, DeepSeek-R1 exhibits notable vulnerabilities regarding content safety and fairness. Recent evaluations show that DeepSeek-R1 is likely to produce harmful outputs, including cybersecurity threats, toxic language, and hazardous instructions related to chemical or biological materials. In addition, DeepSeek-R1 exhibits biased behavior in approximately 83% of political, racial, and gender-related prompts. This behavior not only undermines the model's credibility but also poses serious ethical and legal risks—particularly under frameworks such as the European Union's AI Act and the United States Fair Housing Act[11]. In healthcare, biased outputs could inadvertently exacerbate health disparities,[36] compounding existing inequities among marginalized groups and reducing trust in AI-assisted clinical decision-making[10].

**Adversarial misuse and model exploitation**

DeepSeek-R1 is particularly susceptible to adversarial manipulation. Techniques such as prompt injection, fine-tuning attacks, and jailbreaking can be used to coerce the model into producing deceptive, unsafe, or unethical outputs. These vulnerabilities raise concerns around the model's deployment in high-stakes contexts where reliability and robustness are paramount—such as in legal interpretations, cybersecurity responses, and public health communication[37].

**Privacy, accountability, and governance**

Privacy remains a key challenge, especially when deploying LLMs in sensitive domains like healthcare. Without strong guardrails, such as privacy-preserving model architectures, adversarial testing, red-teaming evaluations, and adaptive content filtering, the risk of leaking sensitive or personally identifiable information increases. Moreover, the lack of clear accountability frameworks creates ambiguity around who bears responsibility for model misuse or harmful outcomes. Developers, institutions, and policymakers must establish mechanisms for oversight, liability, and user redress to ensure responsible use.

**Regulatory and legal compliance**

As LLMs increasingly influence decision-making in regulated industries, compliance with existing legal frameworks is essential[11]. However, many regulatory standards—such as the Health Insurance Portability and Accountability Act (HIPAA) in the U.S.—were not designed with generative AI in mind. This regulatory mismatch introduces legal ambiguity and operational risks when deploying models like DeepSeek-R1 in clinical or legal settings. Stronger oversight, industry-wide auditing standards, and the incorporation of AI-specific provisions into existing regulatory frameworks are needed to mitigate unintended consequences and ensure legal conformity.

**FUTURE DIRECTIONS**

As generative AI technologies mature, the research agenda must shift beyond improving model accuracy and scaling up capabilities. Figure 3 illustrates the future directions of DeepSeek-R1. For DeepSeek-R1, the future lies in developing systems that are not only powerful but also aligned with ethical norms, computationally efficient, and adaptable to real-world applications. This evolution requires a concerted effort from the broader AI ecosystem—including developers, regulators, clinicians, ethicists, and end-users—to ensure that the model is safe, fair, transparent, and socially beneficial.

**Advancing model safety and robustness**

One major area for future research involves enhancing the model's safety and alignment with human values. DeepSeek-R1's vulnerability to adversarial manipulation and its capacity to generate unsafe or inappropriate content highlight the need for robust mitigation strategies. These may include improved adversarial defenses, such as training methods resistant to prompt-based attacks, and the implementation of dynamic moderation systems capable of identifying and filtering harmful outputs in real time.[38] Crucially, these technical safeguards must be complemented by continuous ethical alignment processes—such as incorporating human feedback loops and ethical auditing practices—that ensure the model's behavior remains consistent with societal expectations.

**Mitigating bias and enhancing fairness**

Addressing algorithmic bias remains another high-priority research area. Efforts to reduce bias must begin at the data curation stage, where researchers should apply rigorous preprocessing techniques to eliminate harmful stereotypes and ensure that underrepresented populations are adequately included. During model fine-tuning, algorithmic fairness methods should be applied to detect and correct any residual discriminatory tendencies. These practices are particularly important in domains like healthcare and criminal justice, where biased recommendations could have life-altering consequences.

**Improving computational efficiency**

Token-intensive reasoning and high inference costs can limit the model's accessibility and usability, particularly in mobile or low-bandwidth settings. Future research should explore hybrid reinforcement learning methods and more efficient optimization algorithms that maintain high performance while reducing resource demands. Such innovations would not only lower operational costs but also democratize access to advanced AI capabilities [14].

**Expanding applications in clinical decision support**

The expansion of DeepSeek-R1's clinical applications represents another promising direction[23]. While early studies indicate the model's potential to outperform traditional statistical methods in areas like DDI prediction and personalized medicine, more work is needed to validate its utility across medical specialties.[28] By demonstrating consistent performance in domains such as cardiology, oncology, and psychiatry, DeepSeek-R1 could solidify its role as a trustworthy clinical decision-support tool for both physicians and researchers[28].

**Enhancing natural language understanding and code generation**

Natural language understanding and code generation are additional areas ripe for enhancement. To compete with top-tier models like GPT-4o, DeepSeek-R1 must refine its conversational abilities—achieving greater fluency, coherence, and contextual awareness in open-ended interactions[33]. Simultaneously, its code generation capabilities should be improved to ensure that outputs are secure, reliable, and aligned with best practices in software development and cybersecurity. These improvements will make the model more useful for tasks ranging from educational tutoring to secure software engineering[37].

**Promoting transparency and ethical governance**

Finally, transparency and governance should remain at the forefront of DeepSeek-R1's development roadmap. Publishing detailed documentation on training data, benchmarking protocols, and model performance across diverse settings will foster greater public trust and enable external scrutiny. Independent third-party audits can help verify safety claims and identify unintended harms, while collaboration with policymakers and ethicists can guide the responsible deployment of the model in society. Compliance with emerging regulatory guidelines—particularly those related to fairness, accountability, and transparency—will be critical as AI systems become more deeply integrated into public and private sectors.

**DISCUSSION**

DeepSeek-R1 has established itself as a highly capable and economically viable LLM, demonstrating impressive reasoning capabilities across a diverse range of domains, including clinical decision support, financial forecasting, and mathematical problem solving. Its open-source availability, combined with remarkable cost-efficiency, positions it as a strong alternative to proprietary models such as OpenAI's GPT series or Anthropic's Claude, particularly for institutions and organizations with limited computational resources. However, despite its promising performance profile and affordability, several technical, ethical, and societal challenges must be addressed to ensure its responsible deployment and long-term sustainability.

One of DeepSeek-R1's most notable strengths lies in its proficiency with structured reasoning tasks. Benchmarking results suggest that the model performs on par with, and in many cases surpasses, leading commercial alternatives when handling logic-intensive problems, symbolic manipulation, and algorithmic reasoning.[39] This strength has particular implications for high-stakes fields such as clinical diagnostics, epidemiological modeling, and legal reasoning, where accuracy and transparency are paramount.[40] Moreover, DeepSeek-R1's computational efficiency translates into lower inference costs, enabling broader accessibility for academic institutions, startups, and governments aiming to integrate AI into their workflows.[41]

The model's open-source nature has also catalyzed rapid innovation within the AI research community. Researchers can fine-tune DeepSeek-R1 for specialized tasks, share enhancements, and build upon a shared foundation—driving iterative improvements at a pace that often exceeds proprietary development cycles. This openness encourages greater transparency and community-driven oversight. However, it simultaneously introduces security vulnerabilities. DeepSeek-R1 is demonstrably more susceptible to adversarial prompt injections, jailbreaking, and misuse during fine-tuning.[29] These vulnerabilities expose the model to potential exploitation, especially when deployed without robust safeguards, and illustrate the inherent trade-off between openness and safety. The challenge, then, lies in finding a sustainable equilibrium between democratized access and robust control mechanisms to prevent malicious use.

Ethical considerations further complicate the landscape. Independent evaluations have revealed that DeepSeek-R1 is significantly more likely than its commercial counterparts to generate biased, toxic, or factually incorrect outputs.[39] This tendency raises critical concerns regarding its alignment with ethical standards and legal mandates, particularly in domains such as healthcare, education, criminal justice, and finance. If left unchecked, these biases could lead to discriminatory practices or the dissemination of harmful misinformation. The stakes are particularly high in healthcare, where flawed recommendations could jeopardize patient safety or exacerbate health disparities. Mitigating these ethical risks requires a multipronged approach.[42] First, developers must invest in more effective bias detection and correction techniques, particularly during the pretraining and fine-tuning stages. Second, the development and application of adversarial training and red-teaming protocols should become standard practice to stress-test the model's behavior under edge-case scenarios. Third, enhanced transparency—such as detailed documentation of training datasets, model architecture, and known limitations—can empower users to make informed decisions and reduce the risk of unintended harm. These steps are critical for ensuring regulatory compliance and fostering public trust in open-source AI systems.[43]

Looking ahead, several technical areas warrant continued research. One priority is improving DeepSeek-R1's natural language understanding and contextual reasoning abilities, which still lag behind frontier models like GPT-4o and Gemini 1.5 in complex dialogue management, long-context reasoning, and real-time comprehension. Addressing these gaps may require innovations in multi-modal learning[44], token-efficient architectures (such as Mixture-of-Experts frameworks), and reinforcement learning with human or synthetic feedback. Additionally, expanding DeepSeek-R1's capacity to handle domain-specific tasks, such as those in biomedical research or scientific computing, will increase its utility and relevance to specialized industries.

In healthcare, for example, DeepSeek-R1 has shown early promise in tasks such as DDI prediction and personalized treatment planning.[28] However, further research is necessary to validate its performance across a wider range of medical specialties,[45] clinical populations, and real-world datasets.[41] Such validation would require close collaboration between AI researchers, clinicians, and health informatics experts to ensure the model's outputs are interpretable, evidence-based, and aligned with current clinical practice.[46] Governance frameworks must also evolve in parallel. The proliferation of open-source LLMs demands coordinated action among AI developers, regulators, ethicists, and civil society stakeholders to define appropriate guardrails for deployment. Policy efforts should focus on clarifying accountability, ensuring fairness, mandating explainability, and enabling third-party audits of high-risk applications.[47] These frameworks should be adaptive, allowing for ongoing revision as the capabilities and societal impact of models like DeepSeek-R1 evolve.

Ultimately, the long-term success of DeepSeek-R1 will depend on its ability to transcend its current limitations and mature into a secure, unbiased, and contextually aware AI system. It must not only compete on performance metrics but also uphold ethical integrity and social responsibility. As generative AI continues to reshape the technological landscape, DeepSeek-R1 serves as both a symbol of innovation and a reminder of the complexities that accompany open, scalable intelligence.[42] Its trajectory illustrates that meaningful progress in AI is not measured solely by raw capability, but by how thoughtfully and inclusively that capability is developed, governed, and applied. Through collaborative stewardship, DeepSeek-R1 can fulfill its potential as a force for equitable and sustainable societal benefit.[48]

**CONCLUSION**

DeepSeek-R1 represents a significant leap forward in the development of open-source LLMs, offering powerful capabilities in structured reasoning, domain-specific inference, and cost-efficient deployment. Its architecture—built on MoE, CoT reasoning, and reinforcement learning—enables high performance in mathematics, healthcare, code generation, and scientific domains, positioning it as a compelling alternative to proprietary systems such as GPT-4o and Claude-3 Opus. Yet this advancement is accompanied by complex trade-offs. DeepSeek-R1's openness fosters transparency, accessibility, and collaborative innovation, particularly in under-resourced environments. However, this same openness increases exposure to adversarial fine-tuning, bias amplification, and misuse—especially in sensitive sectors like healthcare and finance. While its technical innovations address long-standing challenges in AI reasoning and adaptability, the model's alignment vulnerabilities and limitations in natural language understanding, safety, and generalizability underscore the ongoing need for robust oversight and ethical governance. To fulfill its potential as a trustworthy and equitable AI system, DeepSeek-R1 must be paired with rigorous safeguards: improved alignment protocols, domain-specific validation, transparent documentation, and active community stewardship. It also demands regulatory clarity and multi-stakeholder cooperation to mitigate harm and enhance benefit. In this regard, DeepSeek-R1 stands not just as a technological milestone but as a bellwether for the future of responsible open-source AI. Its trajectory will help define whether powerful generative models can be both universally accessible and rigorously safe in real-world practice.

**FUNDING**

None.

**CONTRIBUTION STATEMENT**

JY designed the study, contributed to the data analyses, and the writing of the manuscript. SB, JH, and MH contributed to the writing of the manuscript. All authors read and approved the final version of the manuscript.

**CONFLICT OF INTEREST STATEMENT**

None.

**Table 1.** Comparison of key characteristics of DeepSeek-R1 and other recent LLMs.

| | | DeepSeek-R1 | Mixtral | LlaMA 3 | Qwen | GPT-4o | Claude 3 Opus | Gemini |
|---|---|---|---|---|---|---|---|---|
| **ARCHITECTURE** | Mixture of Experts | ✔ | ✔ | ✘ | ✘ | ✘ | ✘ | ✘ |
| | Reinforcement Learning | ✔ | ✘ | ✘ | ✘ | ✔ | ✔ | ✔ |
| | CoT Reasoning | ✔ | ✘ | ✘ | ✘ | ✔ | ✔ | ✔ |
| | Self-reflection phase in training | ✔ | ✘ | ✘ | ✘ | ✘ | ✘ | ✘ |
| | Multimodal input | ✘ | ✘ | ✘ | ✘ | ✔ | ✘ | ✔ |
| **PERFORMANCE** | High math benchmarks | ✔ | ✔ | ✘ | ✘ | ✔ | ✔ | ⚠ |
| | Clinical support accuracy | ✔ | ✘ | ✘ | ✘ | ✔ | ✔ | ⚠ |
| | Real-time fluency | ⚠ | ✘ | ✘ | ✘ | ✔ | ✔ | ✔ |
| | Specialty-domain coverage | ✔ | ✘ | ✘ | ✘ | ✔ | ✔ | ⚠ |
| **TRANSPARENCY AND ACCESS** | Open-source | ✔ | ✔ | ✔ | ✔ | ✘ | ✘ | ✘ |
| | Public training documentation | ✔ | ⚠ | ✘ | ⚠ | ✘ | ✘ | ✘ |
| | Modifiable | ✔ | ✔ | ✔ | ✔ | ✘ | ✘ | ✘ |
| | No license/API restrictions | ✔ | ✔ | ✔ | ✔ | ✔ | ✔ | ✘ |
| **SAFETY, BIAS, AND MISUSE** | Built-in safety techniques | ✘ | ✘ | ✘ | ✘ | ✘ | ✘ | ✘ |
| | Openness about vulnerabilities | ✔ | ✘ | ✘ | ✘ | ⚠ | ✔ | ⚠ |
| | Customizable safety filters | ✔ | ⚠ | ✔ | ✔ | ✘ | ✘ | ✘ |
| **COST AND EFFICIENCY** | Low inference cost | ✔ | ✔ | ✔ | ✔ | ✘ | ✘ | ✘ |
| | Token-efficient interference | ⚠ | ✔ | ✔ | ✔ | ✘ | ✘ | ✘ |
| | Offline deployment | ✔ | ✔ | ✔ | ✔ | ✘ | ✘ | ✘ |
| | Custom distillation | ✔ | ✔ | ✔ | ✔ | ✘ | ✘ | ✘ |

✔ Model has feature
✘ Model does not have feature
⚠ Partial support or insufficient public information

**Table 2.** Summary of selected studies.

| Authors | Country | Domain | Aim of Research | Key Implications |
| --- | --- | --- | --- | --- |
| Alghamdi H, et al.[49] | Saudi Arabia | Healthcare | ● Predict prescribed medications using different LLMs on EHR data | ● DeepSeek-R1 was among the top performers in predicting medications from EHR data, showing strong NLP capability |
| Arrieta A, et al.[37] | Spain | AI Safety | ● Assess and compare safety alignment of DeepSeek-R1 and OpenAI's o3-mini by generating unsafe test inputs using an automated testing tool | ● DeepSeek-R1 produced significantly more unsafe responses than o3-mini<br>● Suggests that DeepSeek-R1 may require additional alignment to enhance safety performance |
| Aydin O, et al.[50] | Turkey | Technology | ● Compare several generative LLMs for academic content creation, analyzing plagiarism rates, AI detection flags, word count, semantic overlap, and readability | ● Each model produced high volumes of text with moderate-to-high plagiarism detection<br>● All were flagged as AI-generated<br>● DeepSeek-R1 offers potential for broader usage but with reliability and originality concerns |
| Boye J, et al.[13] | Sweden | Mathematics | ● Investigate several LLMs' performance on high-school-level math word problems, test for both correctness and CoT reasoning | ● DeepSeek-R1 showed improved final accuracy, but all models showed flawed or shallow logic steps<br>● Advanced LLMs still need targeted improvements for generalizable structured math solutions |
| Chen A, et al.[18] | China | Technology | ● Compare multiple o1-like LLMs on multilingual MT tasks, analyzing performance, inference overhead, translation patterns, and prompt-based setups | ● DeepSeek-R1 surpassed GPT-4o in specific contextless tasks<br>● DeepSeek-R1 exceled in historical/cultural translations but struggled with "rambling" in Chinese-focused tasks<br>● Emphasized that the token-intensive |

| | | | | "reasoning" approach causes higher costs and longer inference times<br>● Suggests the need for balanced trade-offs (speed vs. quality) and improved instruction-following |
|---|---|---|---|---|
| Chen J, et al.[51] | China | Finance | ● Examine whether ChatGPT and DeepSeek can extract news from WSJ to predict stock returns and macroeconomic factors, comparing them with other LLM-based methods | ● ChatGPT showed higher predictive power for market returns<br>● DeepSeek-R1 underperformed in English due to focus on Chinese<br>● Concludes that ChatGPT more effectively processes English-language economic signals |
| Choudhury A, et al. [30] | United States | Healthcare | ● Determine user trust and adoption for DeepSeek in healthcare | ● Trust in DeepSeek-R1 significantly affects user adoption<br>● Ease of use increases trust, while perceptions of risk decrease trust |
| de Paiva L, et al.[23] | Germany | Healthcare | ● Analyze DeepSeek-R1's performance on USMLE style questions compared with other LLMs | ● DeepSeek-R1 showed competitive results versus established models, emphasizing it being open-source and cost-efficient<br>● Highlights potential for broader medical adoption, although further validation is needed |
| De Vito G, et al.[28] | Italy | Healthcare | ● Investigate LLM-based performance for predicting DDIs, comparing several models (including DeepSeek R1 variants) with classical ML baselines | ● Fine-tuned smaller LLM (Phi-3.5 2.7B) outperformed larger models, but some DeepSeek-R1 variants ranked highly<br>● Highlights that cost-effective smaller LLMs can excel in DDI tasks, suggesting potential for resource-efficient pharmaco-informatics solutions |

| Dong B, et al.[52] | United States | Technology | ● Examine DeepSeek R1's performance in TEE setups, examining overhead, and GPU usage | ● Smaller models (1.5B) effectively utilized TEE, surpassing CPU-only performance<br>● Larger models, including DeepSeek R1, saw big overhead<br>● Suggests CPU-GPU synergy and code optimization can unify security and high-speed inference, especially for resource-limited secure cloud settings |
|---|---|---|---|---|
| Evstafev E[14] | United Kingdom | Mathematics | ● Examine DeepSeek-R1's performance on challenging math problems from the MATH dataset, specifically regarding token usage, speed, and accuracy | ● DeepSeek-R1 achieved superior accuracy on difficult math tasks but consumed significantly more tokens than comparable models<br>● Suggests a trade-off between thorough multi-step reasoning and efficiency |
| Fu J, et al.[53] | United Kingdom | Technology | ● Evaluate how LLMs (DeepSeek-V3, DeepSeek-R1, ChatGPT, etc.) can help produce short videos with higher popularity metrics, using "popularity prediction" and prompt enhancements | ● DeepSeek-V3 and -R1 can create content matching or exceeding typical engagement<br>● Token-intense reasoning may slow inference<br>● Prompt engineering significantly improves popularity. |
| Guo D, et al.[5] | China | Technology | ● Introduce DeepSeek-R1 for advanced multi-step reasoning, and incorporate supervised fine-tuning to improve readability and safety | ● Demonstrates that RL alone can create emergent reasoning behaviors but can degrade readability<br>● The multi-stage pipeline, plus distillation to smaller models, offers a blueprint for open-source advanced reasoning LLMs |
| Gupta G et | United | Healthcare | ● Compare DeepSeek-R1 and o3-mini in | ● DeepSeek-R1 needs further training |
| al.[54] | States | | diagnosing chronic conditions | in respiratory diagnostics before clinical use<br>● Reliable in mental health and neurology |
| Gupta R[55] | India | AI Policy | ● Compare advanced LLMs (DeepSeek R1, ChatGPT, LLaMA, etc.) on CoT reasoning, training cost, and biases around politics or censorship | ● DeepSeek-R1 excelled in multilingual tasks but censored Chinese political content<br>● ChatGPT showed broader coverage but different bias patterns.<br>● Each model's dataset shapes bias, calling for improved transparency, data curation, and balanced policy for censorship |
| Krause D[8] | United States | Financial Technology | ● Explore how DeepSeek-R1 is altering traditional development in financial services and fostering innovation, while raising regulatory, ethical, and political questions | ● Lower barriers to entry for startups and enhanced financial inclusion<br>● Highlights serious concerns around security, data governance, and shifting power balances in global AI<br>● Concludes that regulatory frameworks are necessary to harness benefits while mitigating risks |
| Kuo M, et al.[6] | United States | AI Safety | ● Demonstrate a "H-CoT" exploit that rewrites CoT reasoning, bypassing LRM safety checks<br>● Evaluate several LLMs on malicious queries | ● Under a "Malicious-Educator" scenario, refusal rates dropped drastically<br>● Concludes advanced reasoning can be weaponized to produce harmful text<br>● Urges more thorough, consistent safety strategies to preserve beneficial reasoning |
| Liang J, et al. [56] | China | Healthcare | ● Evaluate different LLM performance in oncology-related diagnosis and | ● Reinforces DeepSeek-R1's potential to work alongside physicians in |
| | | | management tasks | ● managing thoracic cancer cases<br>● Strong capabilities in interpreting biological concepts<br>● High accuracy and clarity |
| Lin L, et al. [57] | China | Healthcare | ● Use DeepSeek-R1 to design drug formulas and predict dissolution rates | ● Shows DeepSeek-R1's potential to improve drug formulation workflows using accelerated and accurate prediction models |
| Long Y, et al.[26] | China | Healthcare | ● Develop a no-code, privacy-preserving LLM tool for kidney disease support | ● DeepSeek-R1 enabled high-accuracy, privacy-preserving kidney disease support through a no-code deployment |
| Mikhail D, et al.[17] | Canada | Healthcare | ● Compare DeepSeek-R1 with OpenAI o1 on diagnosing and managing ophthalmic clinical cases, while also examining usage costs | ● Both models reached ~82% accuracy overall.<br>● DeepSeek-R1 offered a 15-fold reduction in token-based costs, indicating a cost-effective alternative in ophthalmic decision support |
| Mondillo G, et al.[22] | Italy | Healthcare | ● Evaluate diagnostic accuracy and clinical utility of ChatGPT-o1 vs. DeepSeek-R1 in pediatric scenarios using MedQA data | ● ChatGPT-o1 outperformed DeepSeek-R1 on pediatric questions<br>● ChatGPT-o1's CoT approach was more structured, while DeepSeek-R1's open-source flexibility makes it valuable for resource-limited settings |
| Neha F, et al.[19] | United States | Technology | ● Offer an overview of DeepSeek's architectures, including R1, and discuss their impact and limitations | ● Highlights specialized MoE frameworks, multi-step reasoning, and cost-effectiveness |
| Parmar M, et al.[20] | India | AI Safety | ● Evaluate limitations of using RL alone to ensure harmlessness in DeepSeek-R1<br>● Compare RL with SFT for alignment and propose hybrid approaches for safer | ● Demonstrates that purely RL-based alignment can suffer from reward hacking, generalization failures, and heavy computational costs |
| | | | deployment | ● Concludes that hybrid RL+SFT approaches are more comprehensive for AI safety<br>● DeepSeek-R1 requires additional strategies to reduce harmful outputs |
| Qui P, et al. [58] | China | Healthcare | ● Evaluate reasoning abilities of LLMs using MedR-Bench | ● DeepSeek-R1 showed strong diagnostic reasoning (>85% in simple cases), but struggled with complex planning tasks |
| Xu P, et al. [59] | China | Healthcare | ● Test DeepSeek-R1's bilingual diagnostic and management reasoning in ophthalmology | ● Supports DeepSeek-R1's adoption in international clinical environments that require complex reasoning across languages |
| Xu Z, et al.[11] | United Kingdom | Computer Security | ● Investigate how fine-tuning attacks compromise safety alignment in DeepSeek-R1, potentially inducing harmful outputs | ● Fine-tuning attacks can significantly undermine CoT-enabled model safety<br>● DeepSeek-R1's advanced reasoning makes it vulnerable when adversarially fine-tuned, stating the need for defense strategies |
| Zhang W, et al[7]. | Chins | AI Safety | ● Evaluate the safety alignment of DeepSeek-R1 and DeepSeek-V3 in Chinese use cases with a new CHiSafetyBench<br>● Reveal model vulnerabilities, particularly with malicious or harmful content | ● DeepSeek-R1 failed to block harmful Chinese prompts in many categories<br>● Concludes that advanced reasoning does not guarantee safety, calling for further alignment<br>● Emphasizes Chinese-specific safety checks for open-source AI |
| Zhan Z, et al. [60] | United States | Healthcare | ● Evaluate DeepSeek models on 4 biomedical NLP tasks | ● DeepSeek-R1 performed well in named entity recognition and classification<br>● Showed limitations in relation/event extraction |

| Zhao M, et al.[25] | China, Australia | Healthcare | ● Assess how several LLMs (ChatGPT variants, DeepSeek-R1 and -V3) generate scoliosis education texts<br>● Measure readability and quality | ● Finds DeepSeek-R1 gives best readability, though overall content quality is similar across models<br>● All models lack citations, which hinders trust<br>● Suggests future LLMs address citations, accessibility, and medical reliability |
|---|---|---|---|---|
| Zhou K, et al.[35] | United States | AI Safety | ● Present a systematic safety evaluation of large reasoning models, including DeepSeek-R1, such as benchmarks and adversarial attacks | ● Identified a large safety gap between open-source R1 and closed-source models<br>● Stronger reasoning can generate more harmful outputs if not well-aligned<br>● Calls for enhanced safety training for open-source reasoning models |
| Zuo Y, et al.[24] | China | Healthcare | ● Develop MedXpertQA for expert-level reasoning and multimodal evaluation for several LLMs | ● DeepSeek-R1 was evaluated and ranked among the top models on complex expert-level reasoning tasks in MedXpertQA |

**Abbreviations:**
LLM: Large Language Model
CoT: Chain-of-Thought
USMLE: United States Medical Licensing Examination
WSJ: Wall Street Journal
DDI: Drug-Drug Interaction
ML: Machine Learning
GPU: Graphic Processing Unit
CPU: Central Processing Unit
TEE: Trusted Execution Environment
H-CoT: Hijacking Chain-of-Thought
LRM: Large Reasoning Model
SFT: Supervised Fine-Tuning

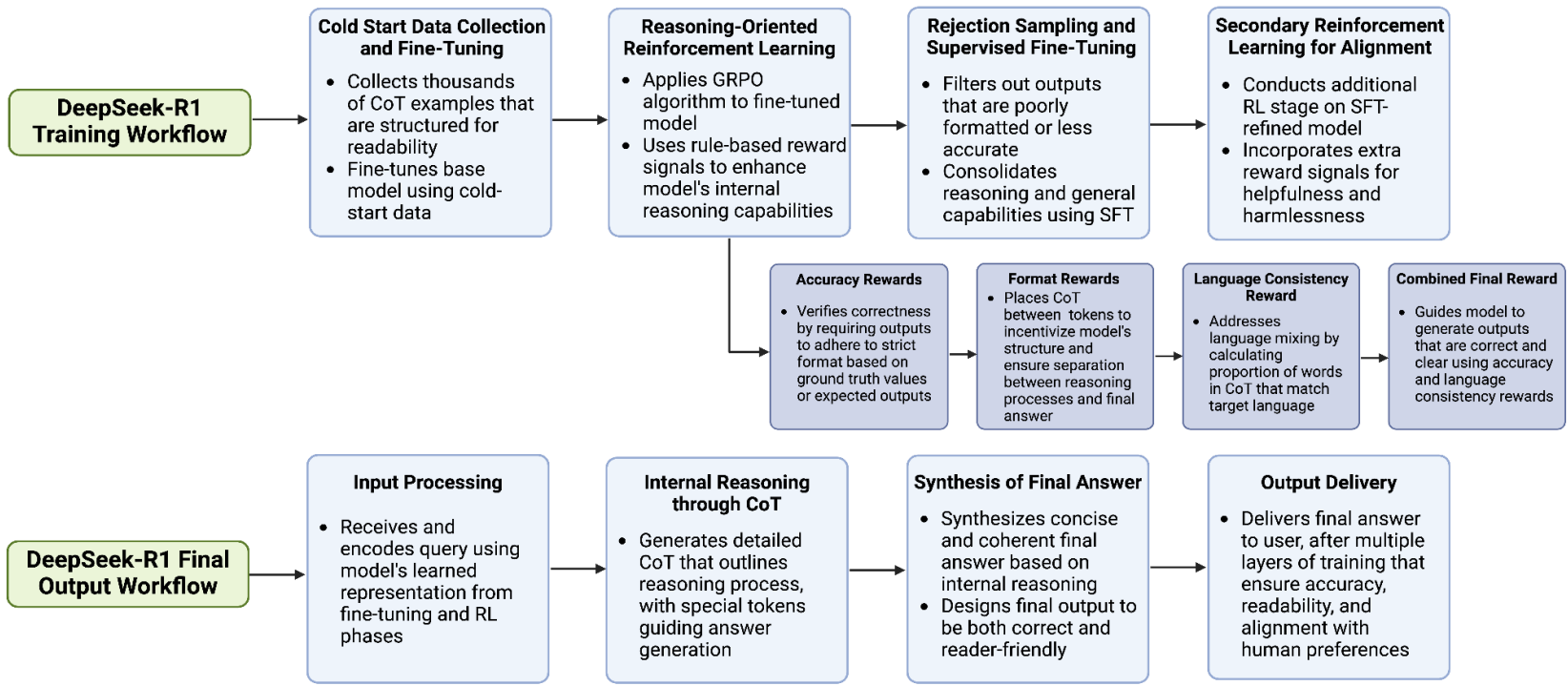

**Figure 1.** Workflow of DeepSeek-R1.

**Abbreviations:**
CoT: Chain-of-Thought
GRPO: Group Relative Policy Optimization
SFT: Supervised Fine Tuning
RL: Reinforcement Learning

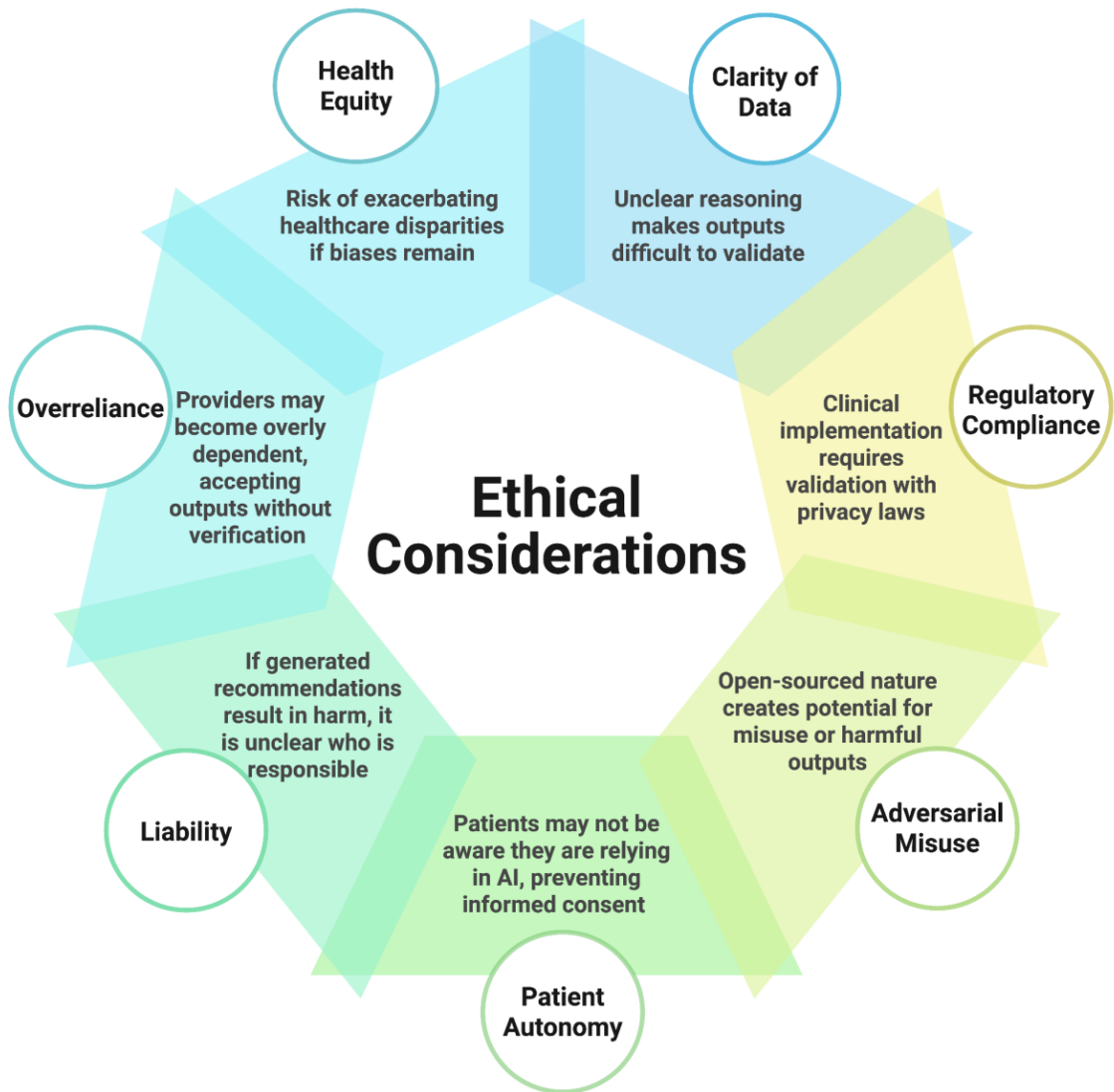

**Figure 2.** Ethical considerations of DeepSeek-R1.

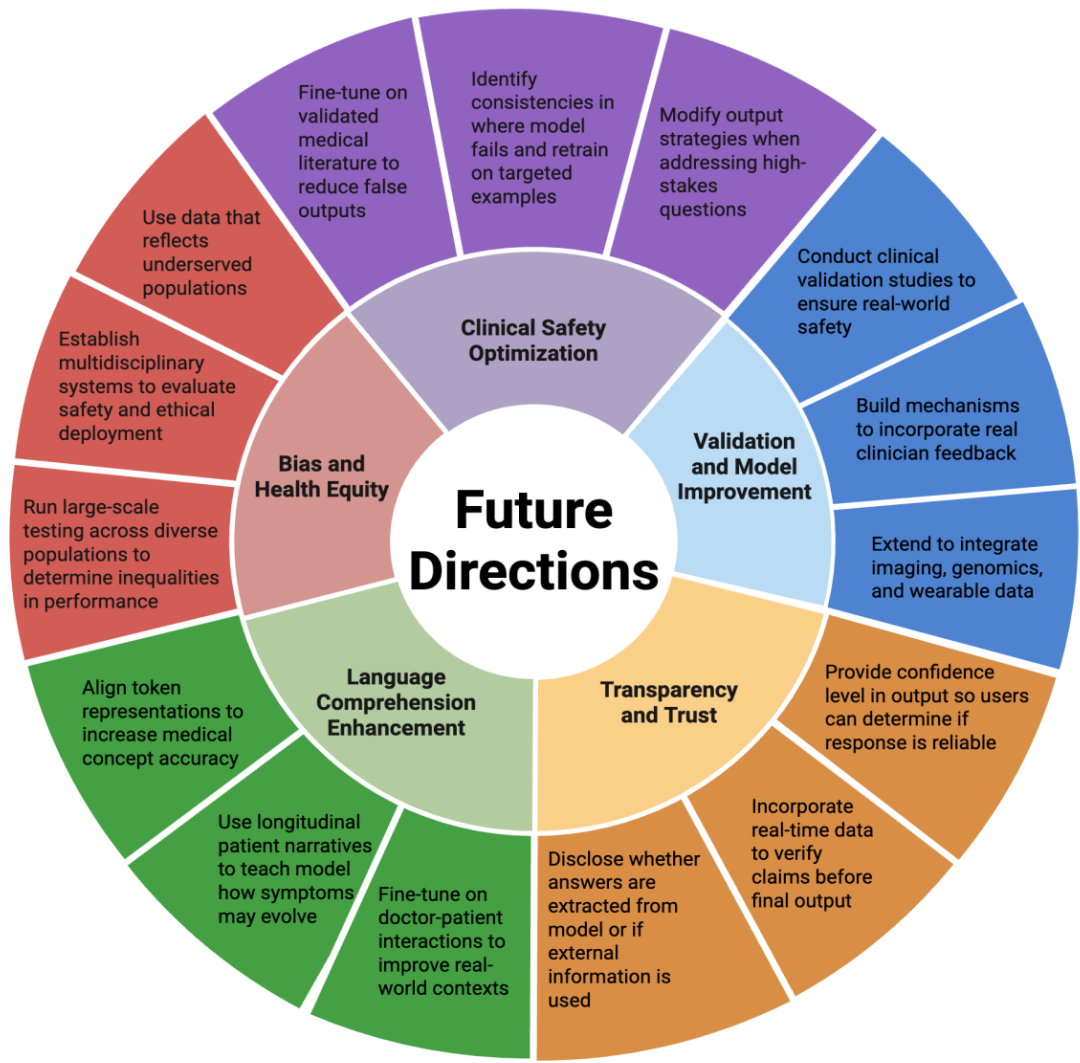

**Figure 3.** Future directions of DeepSeek-R1.